\documentclass[10pt,journal,compsoc]{IEEEtran}
\usepackage{amsmath,amsfonts}
\usepackage{algorithmic}
\usepackage{array}
\usepackage[caption=false,font=normalsize,labelfont=sf,textfont=sf]{subfig}
\usepackage{textcomp}
\usepackage{stfloats}
\usepackage{url}
\usepackage{verbatim}
\usepackage{graphicx}
\usepackage{cite}
\def\BibTeX{{\rm B\kern-.05em{\sc i\kern-.025em b}\kern-.08em
    T\kern-.1667em\lower.7ex\hbox{E}\kern-.125emX}}
\usepackage{balance}
\usepackage{color}
\usepackage{multirow}

\newcommand{\etal}{\textit{et al}.\:}

\title{SpikeCodec: An End–to-end Learned Compression Framework for Spiking Camera}

\author{Kexiang~Feng,
	Chuanmin~Jia,
	Siwei~Ma,~\IEEEmembership{Senior~Member,~IEEE,}
	and~Wen~Gao,~\IEEEmembership{Fellow,~IEEE}
}

\begin{document}	
	\IEEEtitleabstractindextext{%
		\begin{abstract}
			Recently, the bio-inspired spike camera with continuous motion recording capability has attracted tremendous attention due to its ultra high temporal resolution imaging characteristic.
			Such imaging feature results in huge data storage and transmission burden compared to that of traditional camera, raising severe challenge and imminent necessity in compression for spike camera captured content.
			Existing lossy data compression methods could not be applied for compressing spike streams efficiently due to integrate-and-fire characteristic and binarized data structure.
			Considering the imaging principle and information fidelity of spike cameras, we introduce an effective and robust representation of spike streams.
			Based on this representation, we propose a novel learned spike compression framework using scene recovery, variational auto-encoder plus spike simulator.
			To our knowledge, it is the first data-trained model for efficient and robust spike stream compression.
			Extensive experimental results show that our method outperforms the conventional and learning-based codecs, contributing a strong baseline for learned spike data compression.
		\end{abstract}
	
		\begin{IEEEkeywords}
			End-to-end compression, spatiotemporal correlations, attention mechanism, spike compression.
	\end{IEEEkeywords}}

	\maketitle
	\IEEEdisplaynontitleabstractindextext
	\IEEEpeerreviewmaketitle	
	
	\IEEEraisesectionheading{\section{Introduction}}
	\IEEEPARstart{W}{ith} the popularization of autonomous driving, intelligent robotics technology and Industry-4.0 , scene capture and imaging for ultra-high motion dynamics have been increasingly attractive.
	These applications require high temporal resolution for making real-time decisions and adaptive reactions, which is a severe challenge for traditional cameras.
	On the one hand, the shooting target should be static during the time window of rolling shutter scanning, or it will cause motion blur and structure twisting~\cite{meilland2013unified}.
	On the other hand, the imaging facility may fail to record some information or events between two successive shots since its synchronous imaging principle.
	Due to hardware limitations, the imaging speed, measured by frames per second (FPS) of most cameras is less than 1000Hz in general, which means this time window is longer than 1ms between two consecutive snapshot~\cite{schoberl2012photometric}.
	Such characteristic might not satisfy certain requirements for some specific scenarios and applications~\cite{litzenberger2006embedded}~\cite{moeslund2006survey}.
	
	Recently, the neuromorphic imaging cameras were proposed with capability of firing spikes via integrate-and-fire process~\cite{dong2021spike}~\cite{dong2019efficient}.
	This novel product records luminance intensity via firing spikes asynchronously, which breaks the limitation of exposure.
	The luminance can be inferred from different temporal interval between two adjacent spikes, for spikes are fired more frequently in brighter environment compared with in darker one.
	Since the luminance is continuous spatially and temporally, the characteristic of spikes is also correlative.
	Spikes of a given pixel can be predicted implicitly from spikes of adjacent pixels, while spikes at a given time can also be inferred from spikes of previous time~\cite{zhu2019retina}.
	These correlations lead to huge redundancy for storage and transmission, and therefore an efficient compression method for spike data is required urgently.
	
	Compressing spike data directly remains a challenging task, even though there exists correlation within spike streams.
	The major problem is that correlations of spike data are much less salient than those of natural images and videos.
	Spike streams are of binary format, representing whether photon accumulation reaches threshold, which means there is no intermediate value to represent gradual changes of photosensitive state.
	This makes the correlation rarefied and results in inaccurate prediction, both spatially and temporally.
	Besides, the ultra high temporal resolution leads to huge data volume compared with data from traditional cameras, which aggravates difficulties for compression.
	For a spike camera with spatial resolution of 1000$\times$1000 and over 40,000 in temporal domain, the data produced per second manufactures a heavy burden for bandwidth up to 40Gbps.
	Thus high compression ratio is expected urgently.
	Traditional method for image compression can only retain some information with low-frequency, while information in pixel-level will be probably discarded during coding and decoding process~\cite{balle2018variational}~\cite{minnen2018joint}~\cite{cheng2020learned}.
	Above factors all illustrate that previous compression method is unsuitable for spike data.
	
	We argue that there exists a specific type of redundancy, spike triggering redundancy, in spike streams.
	Due to integrate-and-fire model, each pixel maintains an initial state of accumulation, which is unimportant for subsequent spike streams.
	To reduce this redundancy, spikes can first be reconstructed to scenes and finally generated from scenes via integrate-and-fire process~\cite{gerstner2002spiking}~\cite{hodgkin1952quantitative}.
	Fully considering the characteristics of spike streams, we propose an scene reconstruction-based spike compression framework.
	Generally speaking, the proposed framework is composed of three major procedures, which are binary-to-scene reconstruction, scene-to-bit compression and scene-to-binary generation.
	The first part reconstructs scene image from spike streams and the second compresses it with assistance of temporally contextual information, while the third generates spike sequence via integrate-and-fire process.
	With joint optimization, we can represent the content with least distortion using least bits and retain ultra high temporal resolution in the meantime.
	The major contributions of this paper are summarized as follows:
		
	\begin{itemize}
		\item 
		We present the first learned compression framework for continuous spike streams captured by bio-inspired cameras. This framework transforms the binary spike data compression problem into a content-aware scene construction and scene coding problem. As a result, we introduce a novel task, learned spike stream coding, to this community. Importantly, the framework does not impose any restrictions on the choice of codec for scene compression.
		
		\item 
		By leveraging a learning-based codec, we can apply an end-to-end joint optimization mechanism to extract features at different scales. We propose a multi-stage joint optimization scheme to achieve ultra-high temporal resolution spiking stream compression and progressive scene reconstruction. To the best of our knowledge, this is the first end-to-end optimized spike data compression method in the literature.
		
		\item 
		To extract motion characteristics and generate attention maps, we introduce the Spike-Oriented Attention Module (SOAM) and Bi-directional SOAM (BiSOAM). These modules enable the codec to focus more on regions with movement. Extensive experiments on diverse spike sequences demonstrate that our proposed framework achieves state-of-the-art (SOTA) rate-distortion performance, surpassing both conventional and recent learning-based data compression approaches.
	\end{itemize}
	
	\section{Related Work}
	\subsection{Bio-inspired spike camera}
	To meet the urgent demand for data with high FPS, the conception of spike camera has been proposed, which simulates the fovea of retina~\cite{wassle2004parallel}~\cite{masland2012neuronal} and has ultra high temporal resolution more than 20kHz due to asynchronous imaging.
	Different from another outstanding asynchronous camera, called dynamic vision sensor (DVS)~\cite{4444573}~\cite{6889103}~\cite{delbruck2010activity}~\cite{gallego2020event}, spike camera records absolute luminance intensity of each pixel and can provides more visual information.
	Since spikes are fired via integrate-and-fire process, they are fired more frequently in brighter environment compared with in darker one~\cite{zhao2020high}, which can be used for downstream applications.
	
	\subsection{Compression methods for spike data}
	With the increasing volume of images and video, many compression frameworks are proposed in recent years.
	However, previous methods are mostly unsuitable for compressing spike streams.
	The correlation of spikes is much weaker than that of images or videos, which leads to severe challenges for compression.
	Since features with low frequency are preserved preferentially, information in pixel-level will probably be lost during coding and decoding process.
	In addition, high compression ratio is expected urgently due to ultra high temporal resolution.
	In brief, compression methods for spike data should be explored individually.
	
	Recently, spike data coding has been an emerging topic.
	Existing methods concentrate on crafted rule borrowed from conventional hybrid coding.
	Dong \etal proposed a method which divides the spike data into several overlapped voxels~\cite{dong2021spike}.
	However, it decreases the resolution, both in spatial and temporal domain, and can hardly reconstruct the exact location for spikes.
	Bi \etal proposed a strategy of adaptive macro-cube partition based on spike number~\cite{bi2018spike}.
	This method encodes spike location and polarity via different algorithms.
	Further study~\cite{dong2018spike} replaced the partition module with an Octree-based cube partition strategy for more efficient compression.
	Besides, it takes inter-cube prediction into consideration and reaches better performance.
	Zhu \etal designed a specific algorithm for spike streams produced by bio-inspired cameras based on inter spike intervals, which converts compression for discrete data to that for continuous data~\cite{zhu2020hybrid}.
	Spike intervals are calculated as basic data unit and compressed via adaptive polyhedron partition.
	
	\section{Problem Statement}
	\subsection{Problem definition}
	We use \{$I_n$\} to denote real scene sequences and \{$S_n$\} for spike sequences corresponding to \{$I_n$\} , where $n=0,1,2,\dots$ represents frame index.
	Moreover, we use \{$\overline{I_n}$\} to indicate scene sequences reconstructed from \{$S_n$\} and \{$\hat{I_n}$\} for scene sequences decoded from the codec.
	Besides, we use \{$\hat{S_n}$\} to represent spike sequences generated from \{$\overline{I_n}$\} via integrate-and-fire process.
	$S_n,\hat{S_n}\in \{0,1\}^{W\times H}$ are binary format.
	The goal of spike compression is decreasing the code-length of bit streams while keeping the informative fidelity between \{$S_n$\} and \{$\hat{S_n}$\}.
	
	\subsection{Spike generation process}
	The spike camera is composed of two-dimensional photosensitive array.
	Simulating the fovea structure of retina, each position in the array records photon accumulation and fires spikes according to integrate-and-fire procedure~\cite{gerstner2002spiking}~\cite{hodgkin1952quantitative}.
	A spike will be fired when luminance accumulation reaches a threshold, which is referred as integrate-and-fire process.
	This can be formulated as
	\begin{equation}
		\label{if}
		\alpha\int_{0}^{t_0}I_{t}dt \ge \theta,
	\end{equation}
	where $I_{t}$ represents luminance intensity at moment $t$, $\alpha$ and $\theta$ represent photoelectric conversion efficiency and luminance threshold respectively.
	Admittedly, this procedure can be separated into three steps, which are integration, firing and resetting.
	For integration, the hidden stage $\tau_{t_i}$ at moment $t_i$ can be inferred from that at moment $t_{i - 1}$, which is
	\begin{equation}
		\tau_{t_i} = \tau_{t_{i - 1}} + \alpha\int_{t_{i - 1}}^{t_i}I_{t}dt.
	\end{equation}
	For firing, spike $s_{t_i}$ at moment $t_i$ can be decided by $\tau_{t_i}$, which is
	\begin{equation}
		s_{t_i} = \left\{
		\begin{aligned}
			1,&\quad \tau_{t_i} \ge \theta \\
			0,&\quad otherwise
		\end{aligned}
		\right.
		.
	\end{equation}
	$s_{t_i} = 1$ represents a spike is triggered at moment $t_i$, while $s_{t_i} = 0$ for otherwise.
	For resetting, the hidden stage $\tau_{t_i}$ is reset only if $s_{t_i} = 1$.
	Divided by hard and soft scheme, this step can be formulated as
	\begin{equation}
		\tau_{t_i} = \left\{
		\begin{aligned}
			0,&\quad hard\ \ reset\ \ scheme \\
			\tau_{t_i} - \theta,&\quad soft\ \ reset\ \ scheme
		\end{aligned}
		\right.
		.
	\end{equation}
	
	For hardware limitation, pixels can only sample luminance intensity at discrete moments within short period of time, producing temporally discrete spike streams as \{$S_n$\}.
	The signal of $S_n$ denotes the accumulation status at moment $t=nT$, where $T$ is the minimal time interval.
	For simplicity, the spike sequence is divided into several spike planes, each of which represents the firing state of all pixels at a specific moment.
	
	\subsection{Compression-friendly representation of spike streams}
	\begin{figure}
		\centering
		\includegraphics[width=\linewidth]{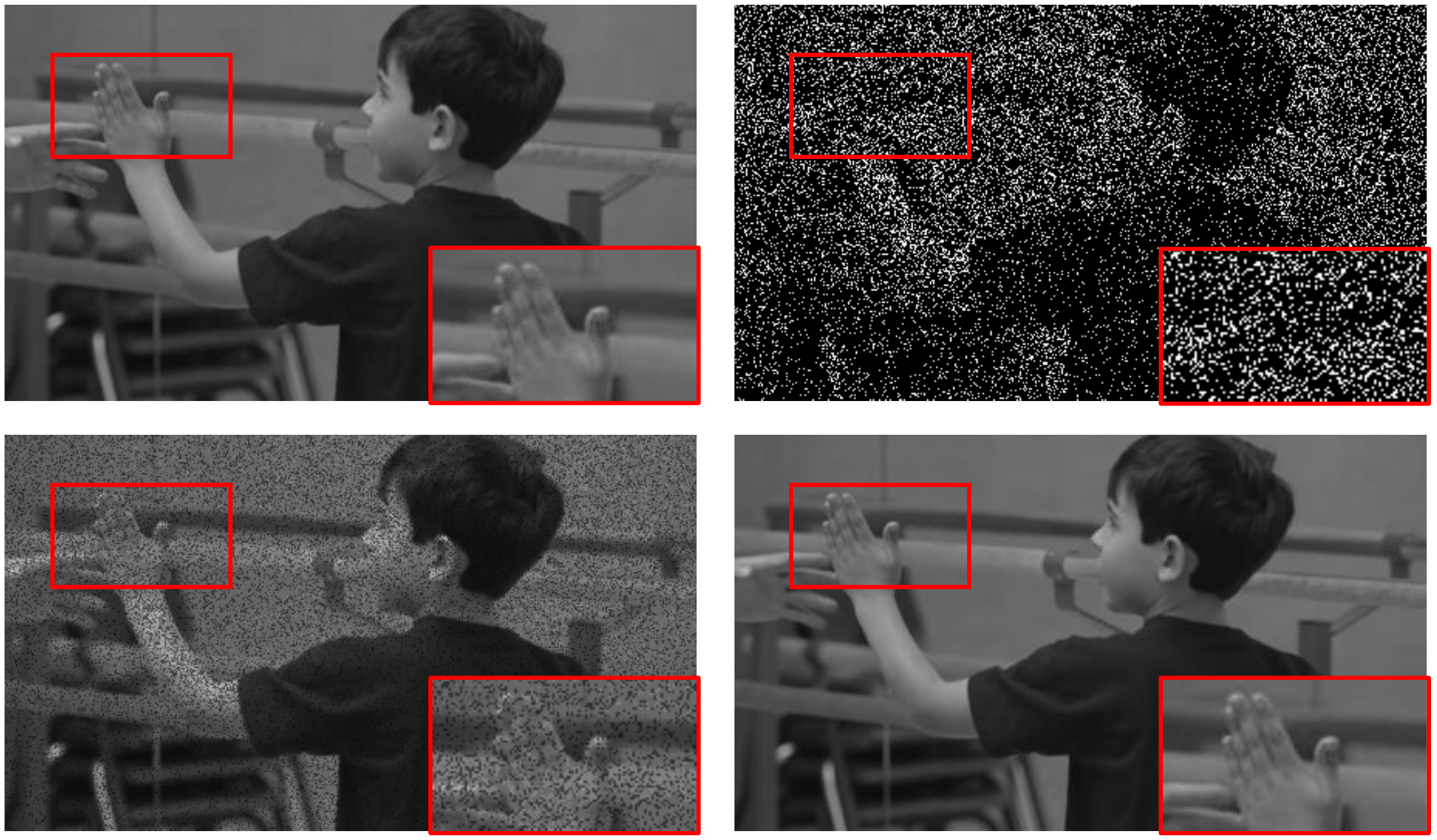}
		\caption{
			Subjective comparison between GT (top left), spike (top right), firing-rate(bottom left) and reconstructed scene (bottom right).
			Spike frame retains little information for its binary format and spatial discontinuity.
			Firing-rate frame (calculated by the reciprocal of ISI) can describe content more precisely, while too much noises exist.
			Compared with first two domain, scene has greater spatial correlation, which is a novel representation of spike data.
		}
		\label{spatial}
	\end{figure}
	\begin{figure}
		\centering
		\includegraphics[width=\linewidth]{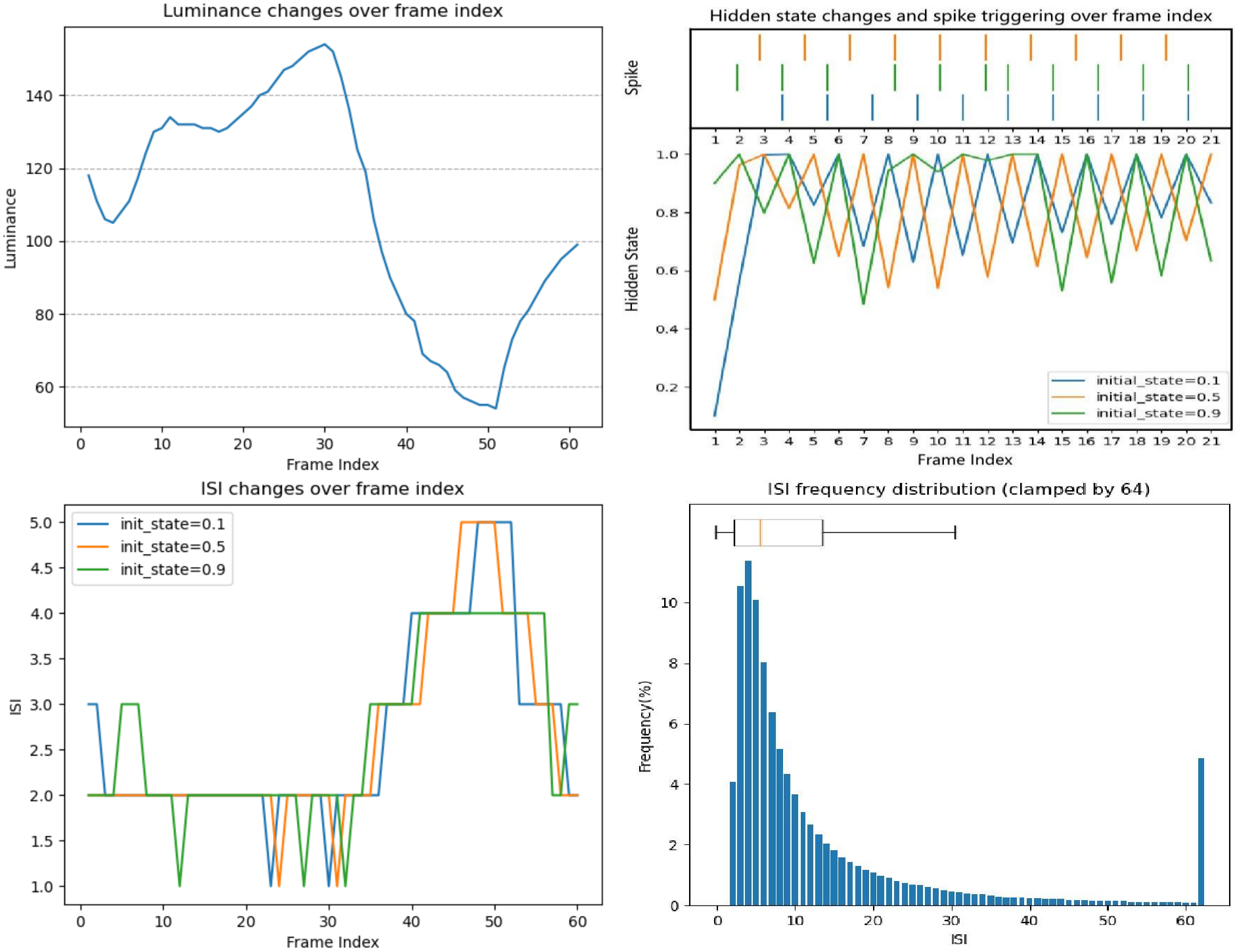}
		\caption{
			Top left: luminance v.s. frame index curve.
			Top right: hidden state and spike v.s. frame index curves with different initial states.
			Bottom left: ISI v.s. frame index curves with different initial states.
			Bottom right: ISI frequency distribution.
		}
		\label{diff_init}
	\end{figure}

	{\bf Spike format \quad}
	Spike generation occurs through an integrate-and-fire process. However, the binary format of spikes can only indicate whether the luminance accumulation reaches the threshold at a specific moment, lacking the ability to represent gradual changes in accumulation. This limitation leads to spatial and temporal discontinuities, which adversely affect inter- and intra-frame correlation, as depicted in Figure~\ref{spatial} (top right panel).
	Spikes can only capture low-frequency content such as the outline of objects like T-shirts and hair, while finer details such as textures in the hand region are difficult to recognize. Furthermore, different initial states can have a significant impact on the spike stream, even when the luminance stimulation is equivalent. In Figure~\ref{diff_init} (top left panel), we compare the integrate-and-fire processes with initial states of 0.1, 0.5, and 0.9, respectively. The top right panel of Figure~\ref{diff_init} illustrates the changes in hidden states over the frame index and the resulting spike triggers.
	It is evident that there are substantial differences in hidden states based on different initial conditions. Similarly, spikes triggered based on hidden states cannot be temporally aligned, resulting in significant variation in spike-wise metrics. This inconsistency is unreasonable since equivalent luminance stimulation should yield spike sequences with consistent characteristics, making spike-based metrics unsuitable for measuring informational fidelity.
	Moreover, identical scene simulations can produce diverse spike streams due to variations in the initial states of the accumulator. All these factors contribute to the challenges associated with directly compressing spike streams.
	
	{\bf ISI format \quad}
	However, we have observed that the inter-spike interval (ISI)\cite{sauer1994reconstruction}\cite{calvin1968synaptic} can effectively overcome these issues. The ISI is calculated as the time difference between two temporally adjacent spikes triggered at moments $t_p$ and $t_n$ ($t_n>t_p$).
	Since there is a one-to-one correspondence between spike and ISI streams, their ability to represent information is strictly equal. By considering the temporal context, the ISI provides a more accurate description of the luminance accumulation process due to its larger value range, which can be considered continuous to some extent. This is illustrated in Figure~\ref{spatial} (bottom left panel), which is reconstructed from the spike firing rate, the reciprocal of the ISI. Both the outline and detailed textures are well restored.
	Furthermore, the ISI maintains higher stability even when the accumulator has random initial states, in comparison to the spike format. Figure~\ref{diff_init} (bottom left panel) demonstrates that ISI changes over the frame index remain relatively consistent regardless of the initial condition. The ISI curves for initial states of 0.1, 0.5, and 0.9 exhibit a similar trend of change, indicating that ISI-based methods are more suitable for measuring informational fidelity.
	Thus, the ISI is an efficient representation for spike data and can be utilized to measure informational fidelity.
	
	Despite the fact that ISI is an equivalent form of spike data, it is not efficient for compression purposes.
	The possible values of ISI are generally distributed within a small range, which limits its ability to describe detailed content effectively. This is evident from the statistical frequency results shown in Figure~\ref{diff_init} (bottom right panel). The box plot indicates that 50\% of ISI values fall within the range of [5, 15], indicating a limited ability to perceive small changes.
	Furthermore, while there is a clear mapping rule between the expected value of ISI and luminance intensity, this relationship is not linear. For instance, assuming a constant luminance value of $I$ during a short period of time, the expected value of ISI can be formulated as follows:
	\begin{equation}
		\label{ISI-luminance}
		\mathbb{E}[{T}^{ISI}_n] = \frac{\theta}{\alpha \mathbb{E}[I_n]}
	\end{equation}
	according to Eq.\eqref{if}, where $\mathbb{E}$ denotes mathematical expectations.
	The differences in luminance intensity are indeed scaled in the ISI domain through injection, which weakens the correlation, especially in the spatial domain. Furthermore, the process of generating ISI is non-robust to noise, as noisy signals are challenging to distinguish from spikes in the binary field.
	Assuming that two temporally adjacent spikes are triggered at $t_p$ and $t_n$ ($t_n>t_p$), noise occurring at $t \in (t_p,t_n)$ will distort the ISI from $t_n-t_p$ to $t-t_p$ and $t_n-t$. This distortion results in poor continuity and significantly impacts the representation of information.
	Therefore, compressing ISI directly can be inefficient.
	
	{\bf Scene format \quad}
	Considering that spikes and ISI are both derived from luminance, they can be inferred from luminance data explicitly or implicitly. We propose that the scene itself can be utilized as a compact and accurate representation of spike data, which is also compression-friendly.
	Compared to ISI, the scene is represented with a deeper bit-depth, allowing for more precise characterization of complex textures. Additionally, the scene exhibits higher spatial correlation, as shown in Figure~\ref{spatial} (bottom right panel), which facilitates efficient hierarchical feature extraction and pixel-wise prediction.
	To quantitatively measure this correlation, we calculate metrics such as variance and conditional entropy, as presented in Table~\ref{quantitative}. For variance, we compute the average value within a region around the target pixel and measure the variance between the predicted and actual values. Similarly, for conditional entropy, we consider the values within the region as conditions and calculate the entropy for all values of the target pixel. The radius of the region is denoted as $r$. The results show that the scene-based metrics achieve better performance, with a variance of \textcolor{red}{\textbf{0.0009}} and a conditional entropy of \textcolor{red}{\textbf{1.3411}}.
	It is important to note that comparing the conditional entropy of spikes is not meaningful, as the binary constraint limits the value range of spike data and increases distortion significantly.
	Furthermore, due to the high correlation in the scene, noises can be easily detected and eliminated using post-processing techniques.
	In conclusion, we believe that the scene itself is a novel and compression-friendly representation of spike data.
	
	\begin{table}
		\centering
		\caption{
			Variance and conditional entropy comparison between spike, ISI and scene formats.
			Data in \textcolor{blue}{blue} should not be compared, for the binary constraint limits value range of spike data and also increases distortion greatly.
			Data in \textcolor{red}{\textbf{red}} denotes scene is a novel representation of spike data.
		}	
		\begin{tabular}{@{}cc|c|c|c@{}}
			\hline\hline
			
			\multicolumn{2}{c|}{\textbf{Metrics}} & \textbf{Spike} & \textbf{ISI} & \textbf{Scene} \\
			\hline
			\multirow{2}{*}{\textbf{Variance} ($\downarrow$)} & $r=1$ & 0.1560 & 0.0038 & \textcolor{red}{\textbf{0.0009}} \\
			\cline{2-5}
			& $r=2$ & 0.1450 & 0.0039 & 0.0018 \\
			\hline
			\multirow{2}{*}{\textbf{Conditional Entropy} ($\downarrow$)} & $r=1$ & \textcolor{blue}{0.6691} & 2.6717 & 1.8329 \\
			\cline{2-5}
			& $r=2$ & \textcolor{blue}{0.6485} & 2.2757 & \textcolor{red}{\textbf{1.3411}} \\
			\hline\hline
		\end{tabular}
		\label{quantitative}
	\end{table}
	
	\section{Methodology}
	\begin{figure*}
		\centering
		\includegraphics[width=\linewidth]{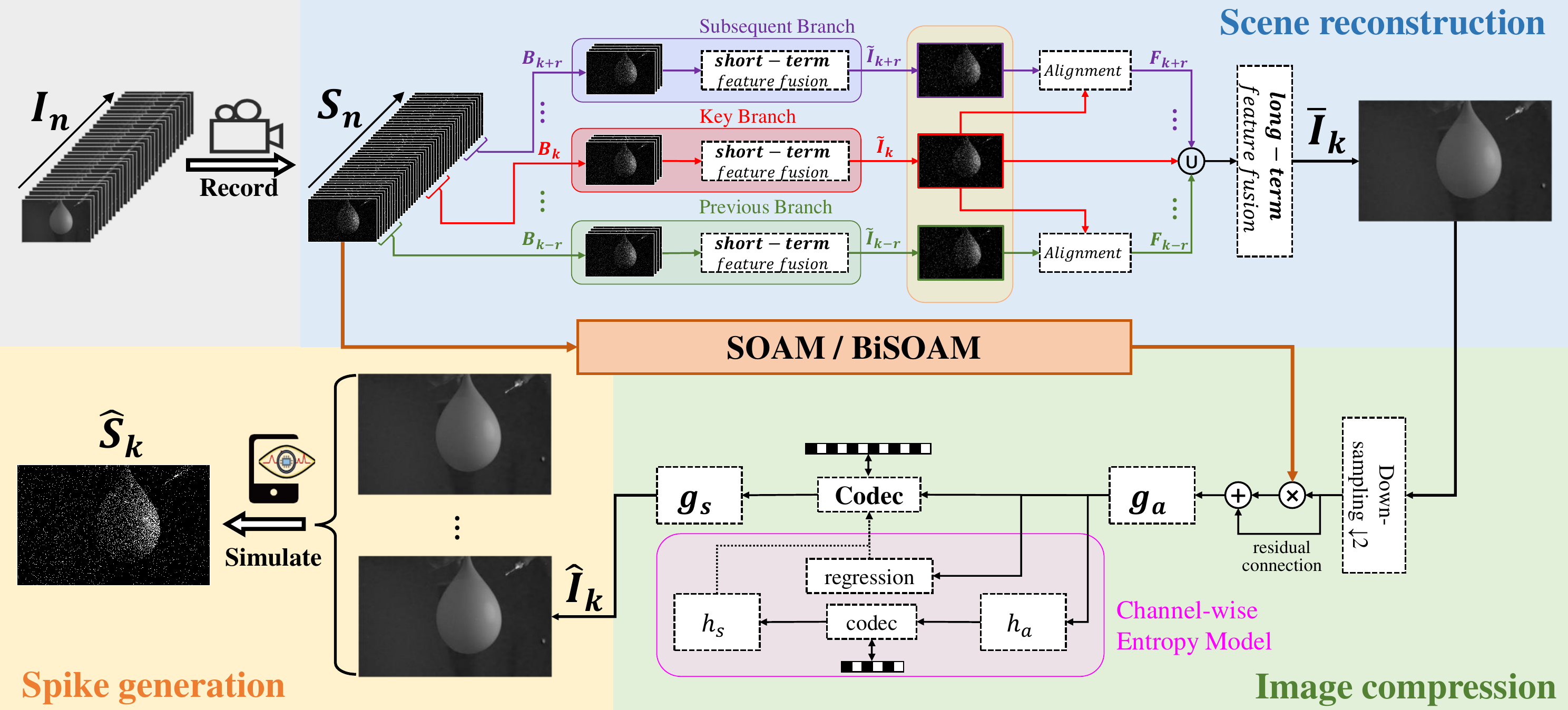}
		\caption{
			Framework of whole spike compression and reconstruction pipeline, which is mainly separated into three steps.
			Blue: Scene reconstruction, which rebuilds scene image from spike stream.
			Green: Image compression, which compresses scene with assistance of temporally contextual information.
			Yellow: Spike generation, which stimulates spike triggering process via integrate-and-fire principle.
			SOAM and BiSOAM are abbreviation of spike oriented attention module and bi-directional spike oriented attention module, which are proposed in section 4.3.
		}
		\label{framework}
	\end{figure*}
	Once we formulate the problem, we propose a novel three-step framework to compress spike sequences efficiently and reconstruct with good informative fidelity.
	In the first step, we convert the continuous spike sequence into images based on long-short term feature extraction.
	In the second step, we compress the images and rebuild with little distortion utilizing information provided by temporal context.
	In the third step, we generate spike sequence from reconstructed images via integrate-and-fire process.
	The first two steps are learned with end-to-end joint optimization scheme, while the third step is not contained for its non-differentiability.
	The overall framework is illustrated in Fig.~\ref{framework} and details of our approach are presented in following subsections.
	
	\subsection{Scene reconstruction module}
	In this module, we reconstruct scene image $\overline{I_k}$ from spike streams ${\{S_n\}}$ using features extracted according to long-short term temporal correlations~\cite{zhao2021spk2imgnet}.
	Because of ultra high temporal resolution and motion continuity, luminance distribution and spike triggering characteristics can be considered as invariant within several frames, which is regarded as short-term correlation.
	Moreover, motion turns violent gradually along with accumulation of frames, representing tendency of movement.
	Both contents and features in pixel-level move follow this tendency, which is considered as long-term correlation.
	Based on long-short term temporal correlations, spike frames can be aggregated to generate a scene image with high quality.
	The module is composed of multi-branches processing and a final fusion operation, which utilizes the thought of divide-and-conquer to extract information from spike stream.
	Specifically, the spike stream is divided into several blocks with overlapping $\{B_t\}$, where $t$ denotes the index of center frame of the block.
	For each $B_t$, spike frames within $\{S_{td - s}, \dots, S_{td + s}\}$ are contained, where $s$ denotes radius of the block and $d$ denotes temporal step length between adjacent blocks.
	$2r + 1$ blocks are taken into consideration totally, representing as $\{B_{t\pm i}\}$ where $i=0, 1, \dots, r$.
	Coarse images $\{\widetilde{I_t}\}$ are then aligned and fused to construct a fine scene image $\overline{I_t}$.

\begin{figure*}
		\centering
		\includegraphics[width=.8\linewidth]{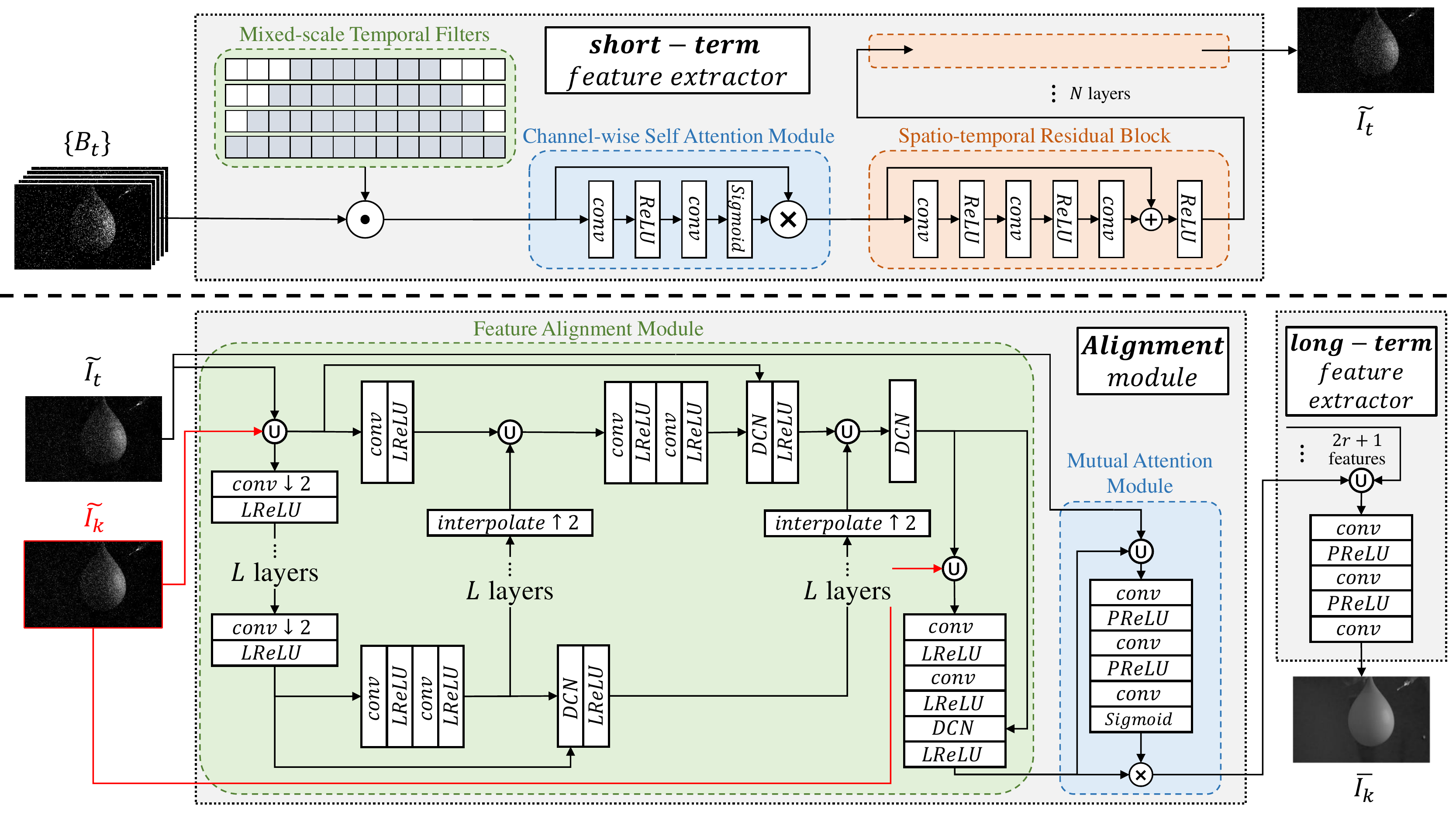}
		\caption{
			Detailed structure of components in scene reconstruction module, including short-term feature extractor (top), alignment module (bottom left) and long-term feature extractor (bottom right).
			With long-short feature extraction, scene can be rebuilt with good quality.
		}
		\label{rec_net}
	\end{figure*}

	Assuming the frame index of target reconstructed scene image is $k$, spike blocks are divided symmetrically, which are $\{B_{k-r},\dots,B_{k+r}\}$.
	For $B_t$ where $t\in[k-r, k+r]$, short-term features are extracted to generate a coarse scene image $\widetilde{I_t}$, which is shown in Fig.~\ref{rec_net} (top panel).
	A set of artificial mixed-scale temporal filters are adopted to obtain information adaptively, followed by a channel-wise self attention module to select appropriate features~\cite{vaswani2017attention}.
	For some scenarios with weaker luminance, photons are generated more tardily, which means it needs more time for pixel to reach threshold and fire spikes.
	In contrast, for some scenarios with high speed motion, only spikes in a short period should be taken into account to avoid blurry reconstruction, since firing rate of spikes can range widely at different moment.
	These illustrate that mixed-scale filters are more suitable to obtain temporal features compared with fixed-scale ones, considering the diversity of motion and environment.
	$N$ layers of spatio-temporal residual block are applied to perceive and further extract features in both spatial and temporal domain, generating preliminary reconstruction image $\widetilde{I_t}$.
	
	After short-term feature extraction, coarse images are aligned and extracted long-term features to produce a refined scene $\overline{I_k}$, which is shown in Fig.~\ref{rec_net} (bottom panel).
	Based on hierarchical strategy, $L$ layers alignment are applied to generate offset maps gradually.
	Maps from bottom layers provide reference for alignment on top layers, resulting in more precise offsets.
	Based on deformable convolutional network (DCN), the alignment module can perceive movements with large scale, increasing the ability of awareness.
	Besides, mutual attention mechanism~\cite{ma2019global} between coarse scene and offset map is employed to generate aligned feature $F_t$.
	Subsequently, $\{F_t\}$ and $\widetilde{I_k}$ are fused and the refined scene $\overline{I_k}$ are rebuilt.
	
	The whole procedure of scene reconstruction module can be formulated as:
	\begin{equation}
		\widetilde{I_t} = \mathcal{STRB}_N[\mathcal{CSA}(\mathcal{F}_{temp}\odot B_t)],
	\end{equation}
	\begin{equation}
		F_t = \mathcal{MA}[\widetilde{I_t}, \mathcal{A}_L(\widetilde{I_k}, \widetilde{I_t})],
	\end{equation}
	\begin{equation}
		\overline{I_k} = \mathbb{F}(F_{k-r}, \dots, F_{k-1},F_{k+1}, \dots, F_{k+r}, \widetilde{I_k}),
	\end{equation}
	where $\mathcal{F}_{temp}$ denotes mixed-scale temporal filters, $CSA$ denotes channel-wise self attention module, $\mathcal{STRB}_N$ denotes spatio-temporal residual block with $N$ layers, $\mathcal{A}_L$ denotes alignment module with $L$ layers, $\mathcal{MA}$ denotes mutual attention module and $\mathbb{F}$ denotes feature fusion and processing.
	
	\subsection{Image compression module}
	In this module, the fine scene image $\overline{I}_k$ is encoded into bit streams, from which a low distorted reconstruction image $\hat{I}_k$ is rebuilt.
	Spatial features are first extracted from image and then be quantified to integers, which is available for calculating probability distribution for entropy coding.
	After being processed by arithmetic codec, features are gathered to reconstruct image with high quality.
	Learning-based codecs assume that features transformed by the encoder conforms to Gaussian distribution.
	Side-information is introduced to minimize Shannon cross entropy of marginal distribution between entropy model and latent~\cite{balle2018variational}.
	Spatial correlation in latent domain is accounted for prediction via auto-regressive module~\cite{minnen2018joint}.
	Besides, attention mechanism is proposed to concentrate more on regions of interest, resulting in allocating code length adaptively and enhance quality of reconstructed images~\cite{cheng2020learned}~\cite{duan2022end}.
	Detailed structure is illustrated in Table.~\ref{com_net}.
	It should be pointed out that the first layer of original $g_a$ is eliminated, for better utilization of the attention map.
	Contrasted  with images of single channel, features of multi-channel can represent information with multi-scale.
	Correspondingly, the attention mechanism can delimit regions with more importance in different scales, which means applying attention mechanism on feature-level brings more informative gain compared with that on pixel-level.
	Thus we remove the first layer out of $g_a$ and utilize attention map on features via residual structure.
	Considering unique characteristics of spike data, we propose a spike oriented attention component , which can utilize information contained in temporal context to guide image compression.
	
	\begin{table}
		\centering
		\scriptsize
		\caption{
			Detailed structure of components in image compression module, including $g_a$, $g_s$, $h_a$ and $h_s$.
			3x3 denotes size of convolutional kernel.
			\textit{c} denotes number of output channel.
			\textit{s} denotes stride of convolution operation, default with 1.
			The first layer of original $g_a$ is eliminated, for better utilization of the attention map.
		}	
		\begin{tabular}{@{}c|c|c|c@{}}
			\hline\hline
			$g_a$ & $g_s$ & $h_a$ & $h_s$\\
			\hline
			TCM & RBU & RBS & RBU \\
			RBS & TCM & TCM & TCM \\
			TCM & RBU & Conv: 3x3 \textit{c}192 \textit{s}2 & Deconv: 3x3 \textit{c}192 \textit{s}2 \\
			RBS & TCM &  & \\
			TCM & RBU &  & \\
			Conv: 3x3 \textit{c}320 \textit{s}2 & TCM &  & \\
			& Deconv: 3x3 \textit{c}1 \textit{s}2 &  & \\
			\hline\hline
			\multicolumn{2}{c}{\textbf{RBS (Residual Block with Stride)}} & \multicolumn{2}{|c}{\textbf{RBU (Residual Block Upsampling)}} \\
			\hline
			\multicolumn{2}{c}{Conv: 3x3 \textit{c}256 \textit{s}2} & \multicolumn{2}{|c}{Deconv: 3x3 \textit{c}256 \textit{s}2} \\
			\multicolumn{2}{c}{Leaky ReLU} & \multicolumn{2}{|c}{Leaky ReLU} \\
			\multicolumn{2}{c}{Conv: 3x3 \textit{c}256} & \multicolumn{2}{|c}{Deconv: 3x3 \textit{c}256} \\
			\multicolumn{2}{c}{Short-cut Addition} & \multicolumn{2}{|c}{Short-cut Addition} \\
			\hline\hline
		\end{tabular}
		\label{com_net}
	\end{table}
	
	\subsection{Spike oriented attention module}
	Due to the significant variation in the distribution of information, allocating equivalent code length for different positions is not cost-effective. Centralization of information arises from two main factors: complex texture and violent motion.
	Texture refers to detailed changes in pixels in the spatial domain~\cite{vaswani2017attention}, while motion represents substantial movement in the temporal domain~\cite{zhu2018end}. Both of these factors require more bits to accurately describe, as neglecting them can lead to blur and distortion.
	To address this issue, an attention mechanism can be employed. The attention mechanism allows for the allocation of diverse levels of attention to different parts of the scene, preserving information according to its importance. In this module, we introduce an attention mechanism to jointly model spatial and temporal information, generating an attention map that guides subsequent compression processes.
	
\begin{figure}
		\centering
		\includegraphics[width=\linewidth]{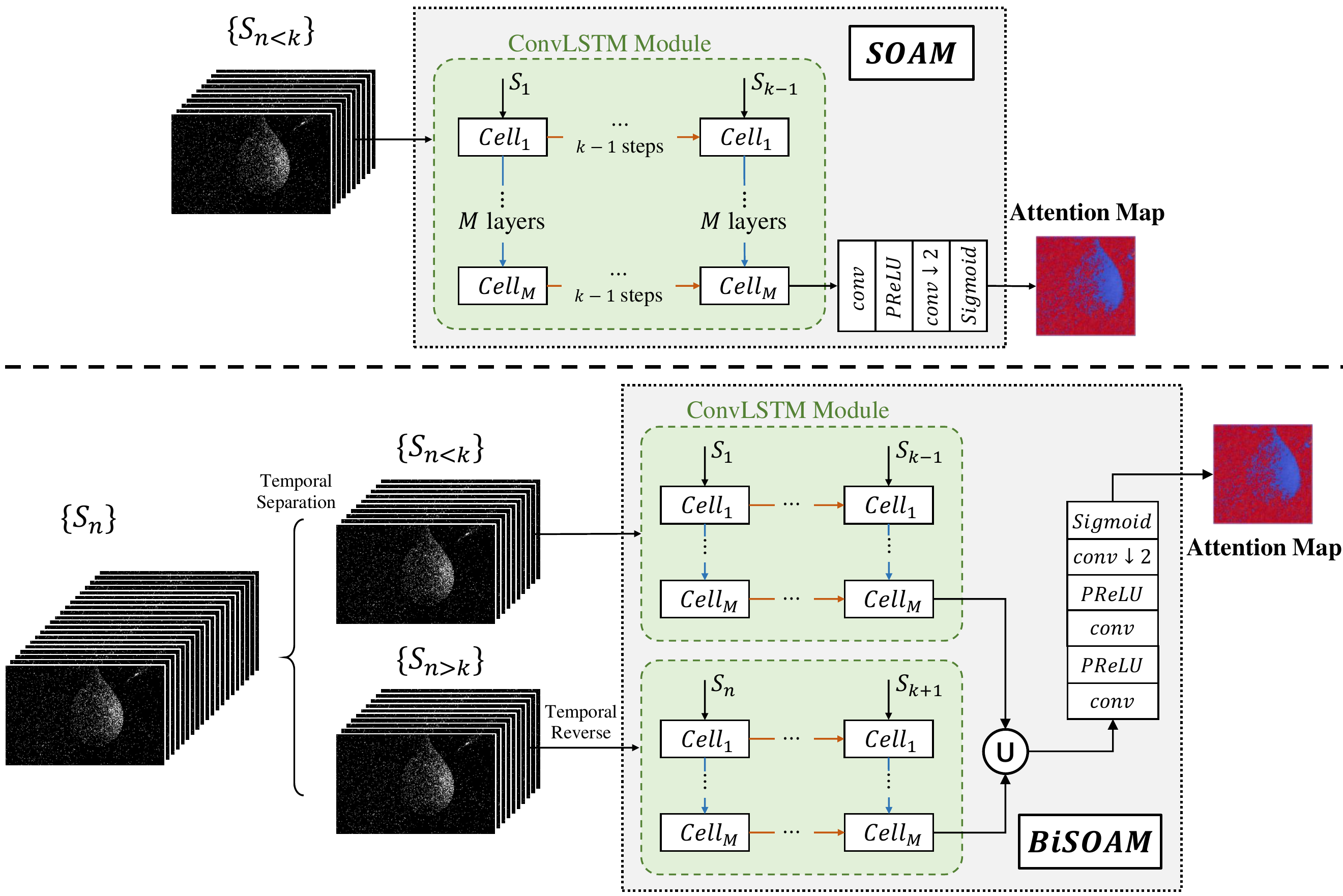}
		\caption{
			Detailed structure of spike oriented attention modules, including SOAM (top) and BiSOAM (bottom).
			With assistant of ConvLSTM module, SOAM and BiSOAM perceive motion nicely and generate the attention map as a guidance for efficient compression.
		}
		\label{soam}
	\end{figure}

	In order to capture the temporal information and incorporate long-term perception, we introduce Long Short Term Memory (LSTM) cells~\cite{sak2014long} for temporal aggregation. Additionally, we propose Spike Oriented Attention Modules (SOAM) based on ConvLSTM~\cite{shi2015convolutional} to efficiently manipulate spatial and temporal information. This architecture is depicted in Figure~\ref{soam} (top panel).
	To leverage the information embedded in continuous motion, we consider a sub-sequence of the spike stream, denoted as $\{S_{n<k}\}$, where $k$ represents the index of the reconstructed frame $\overline{I}_k$. With its ultra-high temporal resolution, the spike stream can describe movement with greater precision, making it suitable for extracting motion characteristics.
	The spike frames are processed by the ConvLSTM module with $M$ layers and $k-1$ time steps. This module not only captures spatial information but also utilizes the temporal continuity of motion. The output of the final cell at the last time step is further utilized to generate an attention map.
	As a result, the attention map effectively delimits the regions of interest, serving as guidance for efficient compression.
	
	Since the motion tendency changes rapidly, the characteristics of $\overline{I}_k$ cannot be predicted solely from the spike sub-sequence before frame index $k$. The sub-sequence after $k$ is also relevant to $\overline{I}_k$ because movement is continuous and reversible. This means that the spike stream $\{S_{n>k}\}$ can also contribute to the aggregation of temporal information.
	Inspired by bi-directional LSTM (BiLSTM)~\cite{huang2015bidirectional}, we propose bi-directional Spike Oriented Attention Modules (BiSOAM) that take both the forward and backward spike sub-sequences into account. The detailed structure of this module is depicted in Figure~\ref{soam} (bottom panel).
	The spike frames $\{S_{n>k}\}$ are also relevant to $\overline{I}_k$ due to the continuous and reversible nature of motion. The entire spike sequence is first divided into two sub-sequences: $\{S_{n<k}\}$ and $\{S_{n>k}\}$. The sub-sequence $\{S_{n>k}\}$ needs to be reversed in the temporal dimension because the motion converges to frame index $k$.
	These two sub-sequences are then processed by ConvLSTM modules, which share parameters. The outputs of the two sub-sequences are fused to produce the attention map. Since the movements contained in $\{S_{n<k}\}$ and $\{S_{n>k}\}$ both converge to $\overline{I}_k$, no alignment operation is required before fusion.
	
	In addition, we have opted not to utilize a convolution-based module for feature extraction in the spatio-temporal domain, despite its exceptional representational capability. The reason for this decision is that it introduces a significant number of parameters and computational complexity exponentially, which is not justified given that features in the temporal domain are comparatively simpler than those in the spatial domain. Moreover, the excessive solution space resulting from convolution-based approaches hinders optimization and makes convergence challenging.
	Furthermore, convolution kernels have a limited ability to perceive long-term information in the temporal domain, as their size is much smaller compared to the temporal dimension. To address these issues, we have chosen to employ LSTM for temporal aggregation. Additionally, we have proposed the SOAM and BiSOAM methods based on ConvLSTM to effectively handle spatial and temporal information, ensuring efficient manipulation and representation.
	
	\section{Experiments}
	\subsection{Datasets}
	Due to page limitation, the detailed training and evaluation protocol are defined in the supplementary material.
	Since there do not exist scene-spike pairs as ground truth (GT), we generate spike sequences from GT scene sequences via simulation methods and regard them associated.
	Vimeo septuplet~\cite{xue2019video} is selected as GT scene sequences, which contains 96 sequences and more than 90k sub-sequences.
	Moreover, we employ the SOTA transformer-CNN~\cite{liu2023learned} as baseline model.
	The proposed SOAM and BiSOAM on the top of baseline model are evaluated.
	In particular, the performance comparisons are provided between baseline, our method, and the latest H.266/VVC reference model (VTM)~\cite{bross2021overview}.
	We illustrate the effecitveness of our framework via two different perspectives, reconstructed image domain and recovered spike data (ISI) domain.
	
	\subsection{Training Details}
	We declare some of our experimental settings for reproducible researches.
	The framework is trained on NVIDIA RTX3090 GPU with CUDA 10.1 and Torch 1.12.
	Initial learning rate is set as $10^{-4}$, and mini batch is 2 due to limitation of memory.
	Rate-distortion function (RD function) is applied for regression~\cite{sullivan1998rate}, in which mean squared error(MSE) is chosen as distortion function.
	Balance parameter $\lambda$ in RD function are set as 0.05, 0.025, 0.013 and 0.0067, which are equivalent to baseline~\cite{liu2023learned}.
	
	As for spike block division, radius of the block is set as $s=6$, temporal step length between adjacent blocks is set as $d=7$ and radius of branches is set as $r=2$.
	As for model parameters, layer of spatio-temporal residual block is set as $N=10$, layer of feature alignment module is set as $L=3$ and layer of ConvLSTM module is set as $M=5$.
	
	The training of our framework involves a multi-stage strategy to address the challenges posed by the complexity of the sub-modules and to facilitate model convergence.
	
	\textbf{Stage I:}
	In this stage, the scene reconstruction module and image compression module are pre-trained separately. These two tasks have somewhat conflicting solution spaces when gradient descent is applied. The scene reconstruction module aims to reconstruct $\overline{I}_k$ to be as consistent with $I_k$ as possible, focusing on capturing motion information from spike streams. The loss function for this module is the distortion between $I_k$ and $\overline{I}_k$, denoted as 
	\begin{equation}
	 	\label{d}
	 	Loss_D=\mathcal{D}(I_k,\overline{I_k})
	 	,
	\end{equation}
	where $\mathcal{D}$ represents a distortion metric such as Mean Squared Error (MSE) or Structural Similarity Index Measure (SSIM).
	On the other hand, the image compression module seeks a balance between bit length and decoded distortion under a specific condition $\lambda$. The loss function for this module is defined as 
	\begin{equation}
		\label{rd}
		Loss_{RD}=\mathcal{R}(y) + \lambda\mathcal{D}(I_k, \hat{I}_k)
		,
	\end{equation}
	where $\mathcal{R}$ estimates the average bit length of the latent distribution $y$, and $\hat{I}_k$ is the reconstructed image from the compressed representation.
	
	\textbf{Stage II:}
	In this stage, joint optimization is performed for the pre-trained scene reconstruction and image compression modules. All parameters in both modules are set as trainable to explore a larger solution space. The loss function for this stage, denoted as $Loss_{RDD}$, includes both the reconstruction loss $Loss_D$ and the compression loss $Loss_{RD}$, with the aim of achieving a balanced optimization:
	\begin{equation}
		\label{rdd}
		Loss_{RDD}=\mathcal{R}(y) + \lambda[\mathcal{D}(I_k, \overline{I}_k) + \mathcal{D}(I_k, \hat{I}_k)]
		.
	\end{equation}
	
	\textbf{Stage III:}
	 In this final stage, the Spike Oriented Attention Modules (SOAM) and Bi-directional Spike Oriented Attention Modules (BiSOAM) are integrated into the network. These modules are trained individually while keeping the parameters of other modules fixed. After convergence, all parameters are set as trainable with a small learning rate. The loss function used in this stage remains the same as $Loss_{RD}$ from Stage I, as the intermediate result $\overline{I}_k$ is expected to have minimal distortion under the small learning rate.
	
	By following this multi-stage training strategy, we aim to progressively optimize the entire model, addressing the challenges introduced by the sub-modules and facilitating convergence to an optimal solution.
	
	\subsection{Compression efficiency in scene domain}
	\begin{figure*}
		\centering
		\includegraphics[width=.9\linewidth]{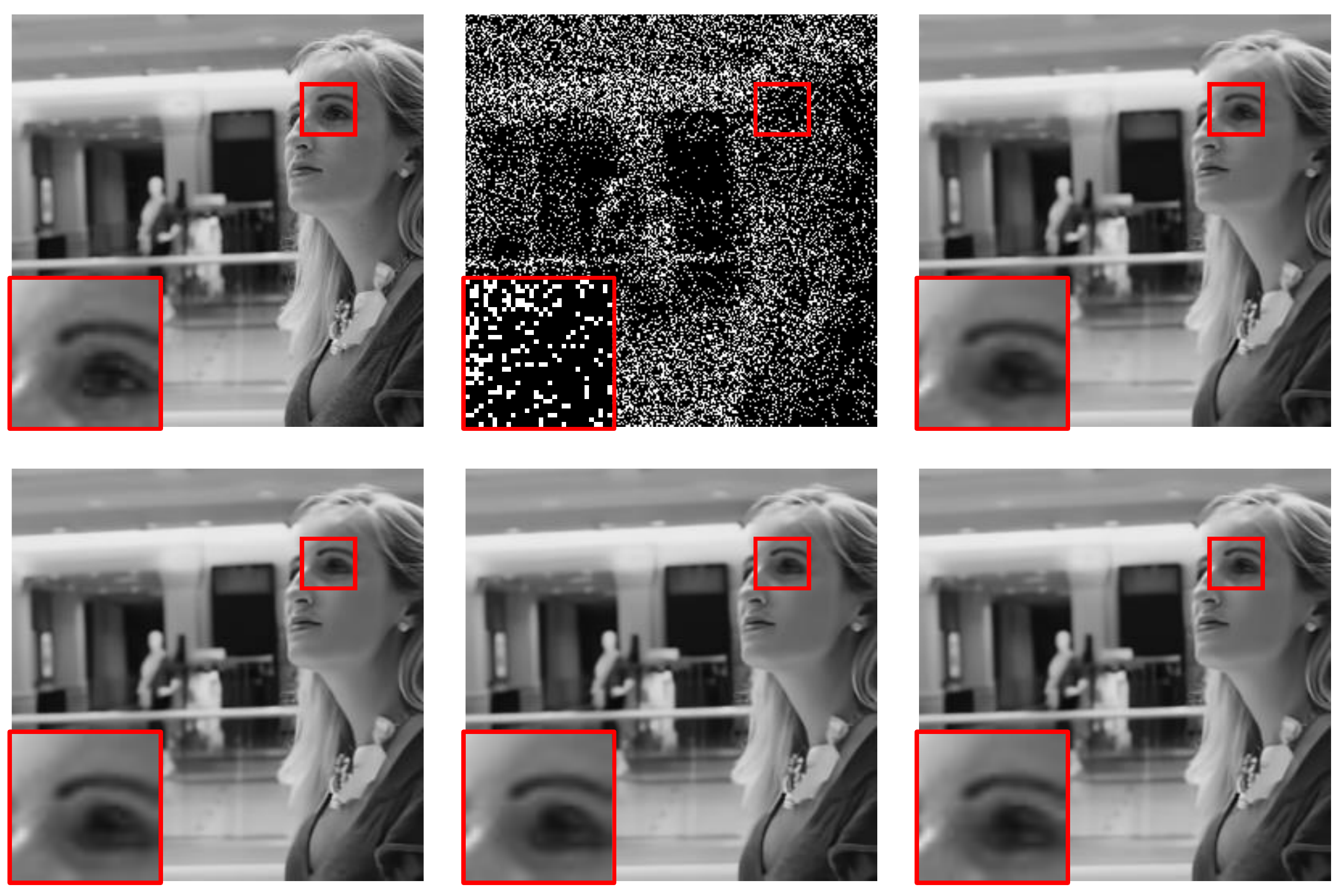}
		\caption{
			Subjective quality of scene reconstruction and compression. 
			Top left: GT scene frame. 
			Top middle: GT spike frame. 
			Top right Reconstruction scene frame via VTM (BPP=0.35,PSNR=36.42dB).
			Bottom left: Reconstruction scene frame via Baseline (BPP=0.33,PSNR=37.29dB).
			Bottom middle: Reconstruction scene frame via Baseline with SOAM (BPP=0.33,PSNR=37.35dB).
			Bottom right: Reconstruction scene frame via Baseline with    BiSOAM (BPP=0.34,PSNR=37.44dB).
			Results show our proposed method outperforms conventional and learning-based codecs, reaching SOTA performance.
		}
		\label{subjective}
	\end{figure*}
	The test set used for evaluation consists of two parts: simulated spike data and real-world captured spike data. For the simulated spike data, we selected real scene sequences from the Vimeo septuplet dataset and generated spike sequences using the integrate-and-fire principle. Since we have access to both the scene sequences and the spike sequences as ground truth, we were able to measure full-reference metrics for performance evaluation.
	In addition to the simulated spike data, we also included spike sequences captured by spike cameras in real-world scenarios. However, since there are no corresponding ground truth scene sequences available for this data, we relied on non-reference metrics for evaluating the performance.
	To assess the effectiveness of our method, we compared it against state-of-the-art learning-based and conventional codecs on both types of data. We presented representative subjective results in Fig.\ref{subjective}, which clearly demonstrate the improved subjective quality achieved by our approach. Furthermore, when comparing at similar bit-rates (0.34 bpp vs. 0.35 bpp), our method outperformed the VTM codec, achieving an increase of nearly 1dB in PSNR (37.44dB vs. 36.42dB).
	
\begin{figure*}
		\centering
		\includegraphics[width=.7\linewidth]{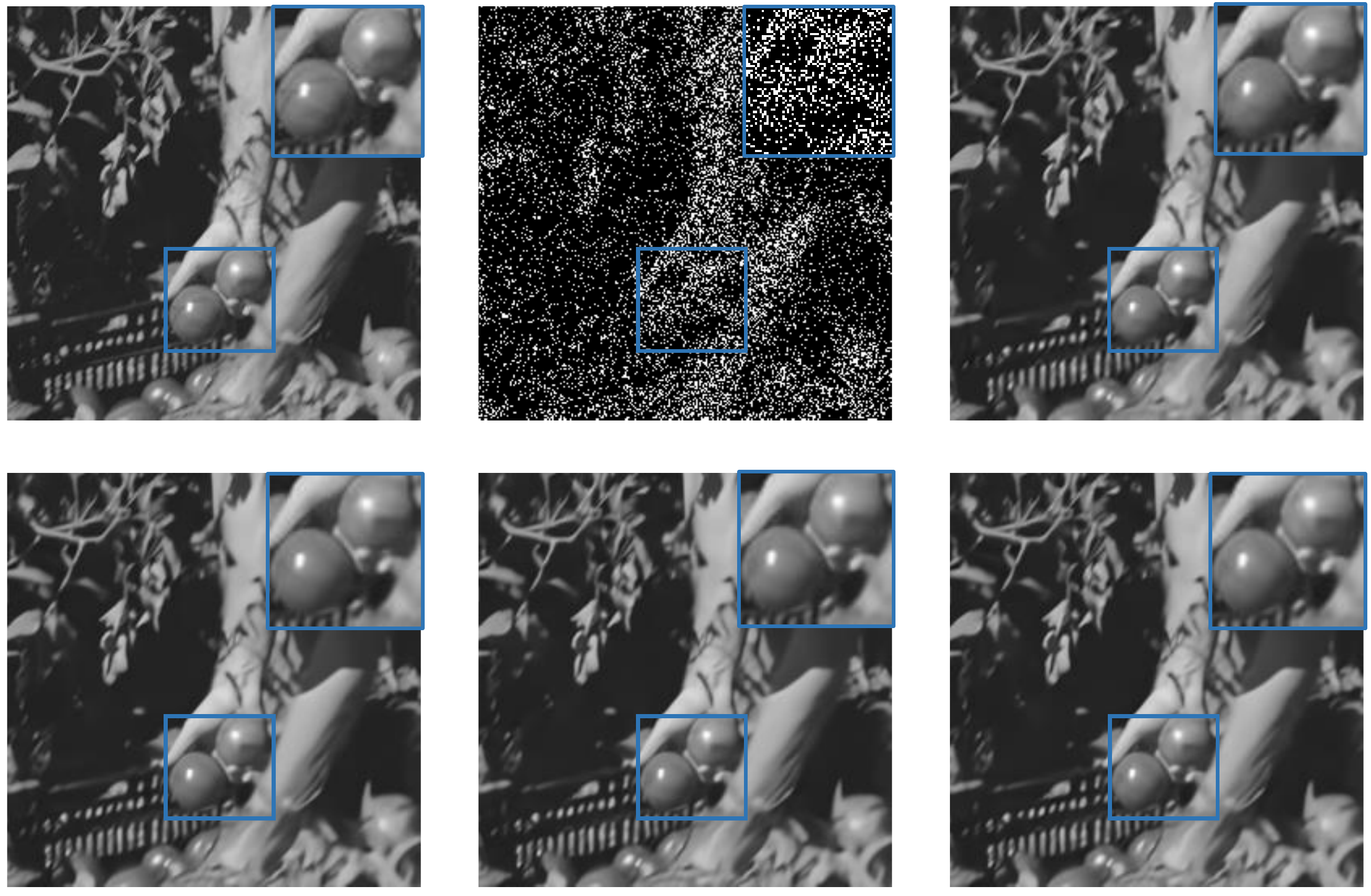}
		\caption{
			Subjective quality of scene reconstruction for low bit-rate ($qp=32$ and $\lambda=0.013$).
			Top left: GT scene frame.
			Top middle: GT spike frame.
			Top left: GT scene frame. 
			Top middle: GT spike frame. 
			Top right: Reconstruction scene frame via VTM (BPP=0.336,PSNR=33.05dB).
			Bottom left: Reconstruction scene frame via Baseline (BPP=0.300,PSNR=33.13dB).
			Bottom middle: Reconstruction scene frame via Baseline with SOAM (BPP=0.300,PSNR=33.25dB).
			Bottom right: Reconstruction scene frame via Baseline with BiSOAM (BPP=0.303,PSNR=33.23dB).
			Results show our proposed method outperforms conventional and learning-based codecs, reaching SOTA performance.
		}
		\label{subjective1}
	\end{figure*}

\begin{figure*}
		\centering
		\includegraphics[width=.7\linewidth]{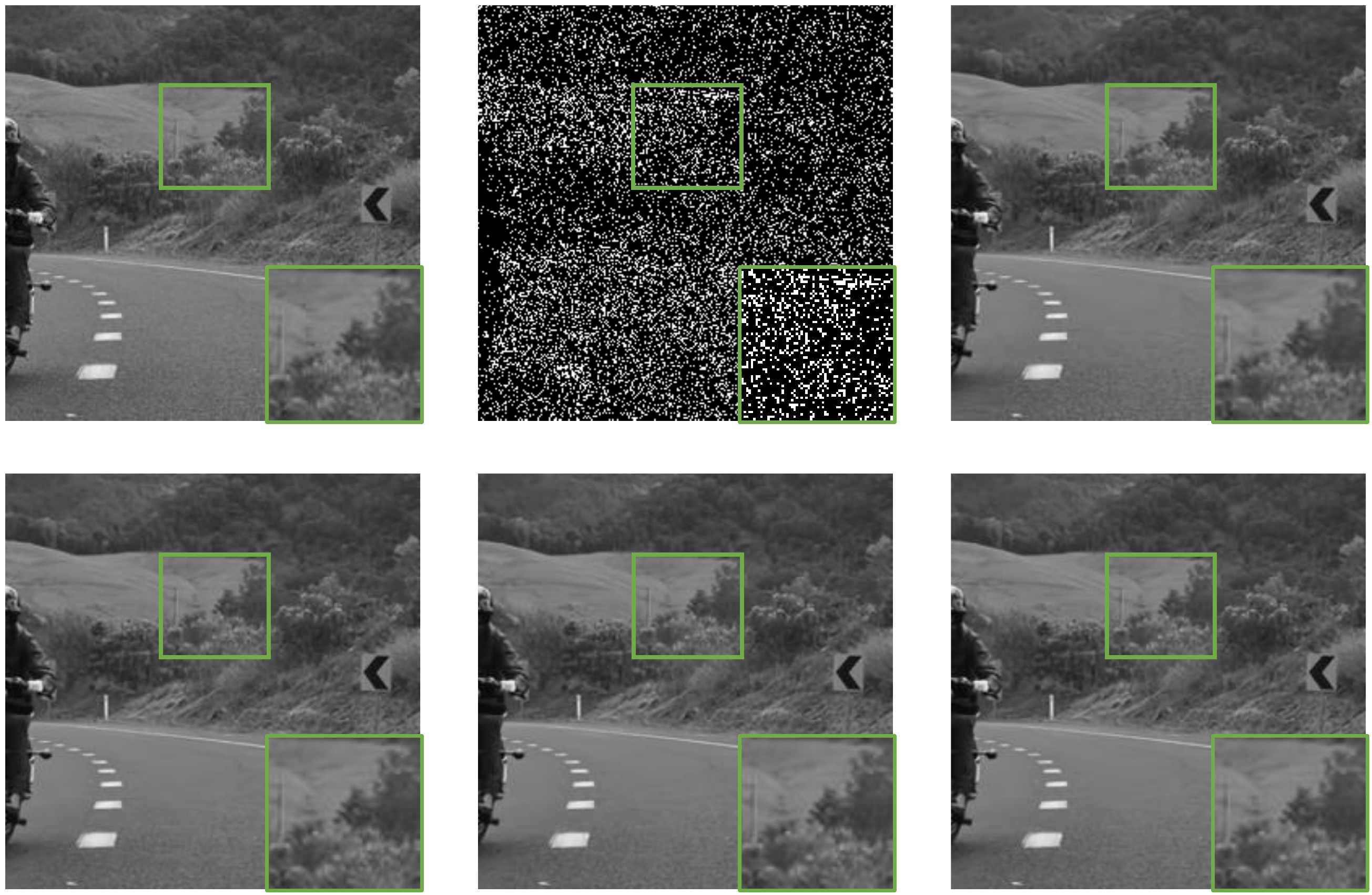}
		\caption{
			Subjective quality of scene reconstruction for high bit-rate ($qp=27$ and $\lambda=0.05$).
			Top left: GT scene frame.
			Top middle: GT spike frame.
			Top left: GT scene frame. 
			Top middle: GT spike frame. 
			Top right: Reconstruction scene frame via VTM (BPP=0.703,PSNR=34.02dB).
			Bottom left: Reconstruction scene frame via Baseline (BPP=0.676,PSNR=35.23dB).
			Bottom middle: Reconstruction scene frame via Baseline with SOAM (BPP=0.677,PSNR=35.24dB).
			Bottom right: Reconstruction scene frame via Baseline with BiSOAM (BPP=0.676,PSNR=33.26dB).
			Results show our proposed method outperforms conventional and learning-based codecs, reaching SOTA performance.
		}
		\label{subjective2}
	\end{figure*}

	To provide a comprehensive analysis, we conducted extensive experiments considering both high and low bit-rate scenarios. For the low bit-rate model, we used a quantization parameter ($qp$) of 32 and a balance parameter ($\lambda$) of 0.013. The results for this case are shown in Fig.\ref{subjective1}. Similarly, for the high bit-rate model, we used $qp=27$ and $\lambda=0.05$, and the corresponding results are presented in Fig.\ref{subjective2}. Our proposed method consistently outperformed both conventional and learning-based codecs across various bit-rates, establishing a strong baseline for learned spike data compression.
	In summary, the experimental results validate the superior performance of our method compared to existing codecs, demonstrating its effectiveness in compressing spike data and enhancing compression quality.
	
	
\begin{figure*}
		\centering
		\includegraphics[width=\linewidth]{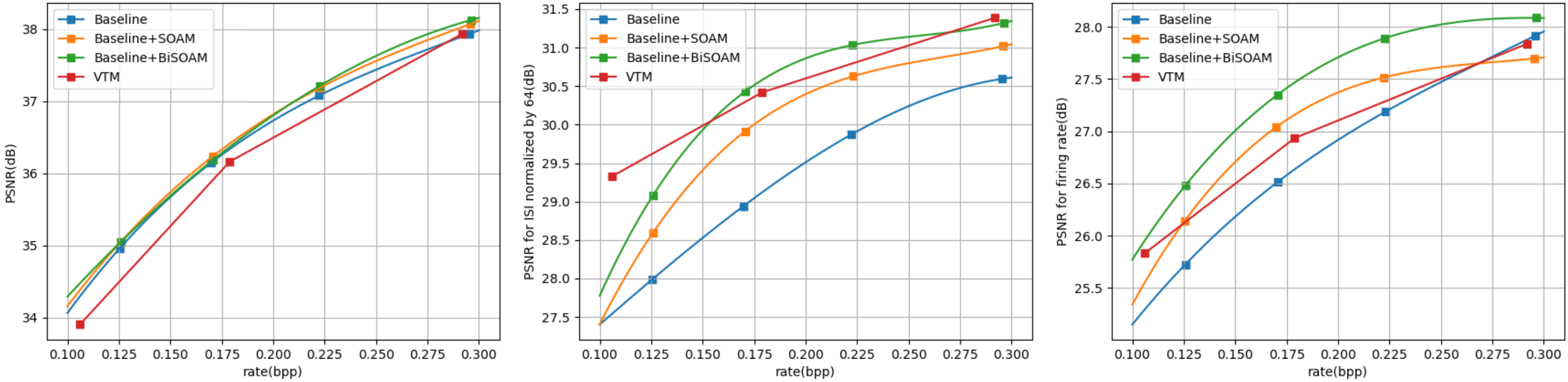}
		\caption{
			PSNR metric for reconstructed image $\hat{I_k}$ (left), for ISI (middle) and firing rate (right) of reconstructed spike sequence $\{\hat{S_n}\}$ v.s. bpp curves of proposed method and baseline.
			Performance of our approach exceeds conventional and learning-based codecs significantly in all domains.
		}
		\label{psnr}
	\end{figure*}

	{\bf Full reference metric based coding performance\quad}
	\begin{table}
		\centering
		\footnotesize
		\caption{
			BD-rate~\cite{bjontegaard2001calculation} gain between VTM, baseline, baseline+SOAM and baseline+BiSOAM.
			Values in table denote gain of model corresponding to row compared with that corresponding to column.
			Negative values represent gain while positive values represent loss.
			Results show that our proposed method outperforms SOTA conventional and learned codecs significantly.
		}	
		\begin{tabular}{@{}c|ccc@{}}
			\hline
			\rule{0pt}{12pt}
			& \textbf{VTM}~\cite{bross2021overview} & \textbf{baseline}~\cite{liu2023learned} & \textbf{baseline + SOAM} \\
			\hline
			\textbf{baseline}~\cite{liu2023learned} & -3.81\% & & \\
			\textbf{baseline + SOAM} & -5.93\% & -2.31\% &  \\
			\textbf{baseline +   BiSOAM} & \textbf{\textcolor{red}{-6.14\%}} & -2.53\% & -0.13\% \\
			\hline
		\end{tabular}
		\label{bdr}
	\end{table}
	In order to assess the objective compression efficiency, we conducted a category experiment and evaluated various metrics. The left panel of Fig.~\ref{psnr} shows the Peak-Signal-Noise-Ratio (PSNR) of the reconstructed image $\hat{I}_k$, as well as the Inter-Spike Interval (ISI) and firing rate of the reconstructed spike sequence ${\hat{S}_n}$. It is evident from the curves that our proposed method outperforms other models in terms of coding efficiency.
	To quantify the improvement, we calculated the BD-rate gain for PSNR compared to the VTM codec, and the average gain was found to be 6.14\%, as shown in Table \ref{bdr}. The rate-distortion (R-D) curves of PSNR metric, presented in various domains, consistently demonstrate the superior coding efficiency of our method for both high and low bit-rate encoding scenarios.
	
	Furthermore, the integration of our proposed SOAM and BiSOAM modules contributes to the rate-distortion performance of the codec, with BD-rate gains of 2.31\% and 2.53\% respectively. Interestingly, we observed that the gain achieved by SOAM is greater than that of BiSOAM for low bit-rate scenarios, while the opposite is true for high bit-rate scenarios. This can be attributed to the fact that at low bit-rates, transporting features with low frequency becomes more crucial in reconstructing scenes, thus favoring the SOAM module. As the bit-rate increases, the availability of sparse bits allows for the transmission of motion information extracted from temporal content, which aids in reconstructing scenes with complex motion. The training process considers the trade-off between motion content and bit-rate, resulting in the observed difference in performance between low and high bit-rate coding scenarios.
	Regarding computational complexity, the end-to-end encoding time for the baseline model is 204ms, while for our model with SOAM and BiSOAM, the times are 211ms and 216ms respectively.
	
\begin{figure*}
		\centering
		\includegraphics[width=\linewidth]{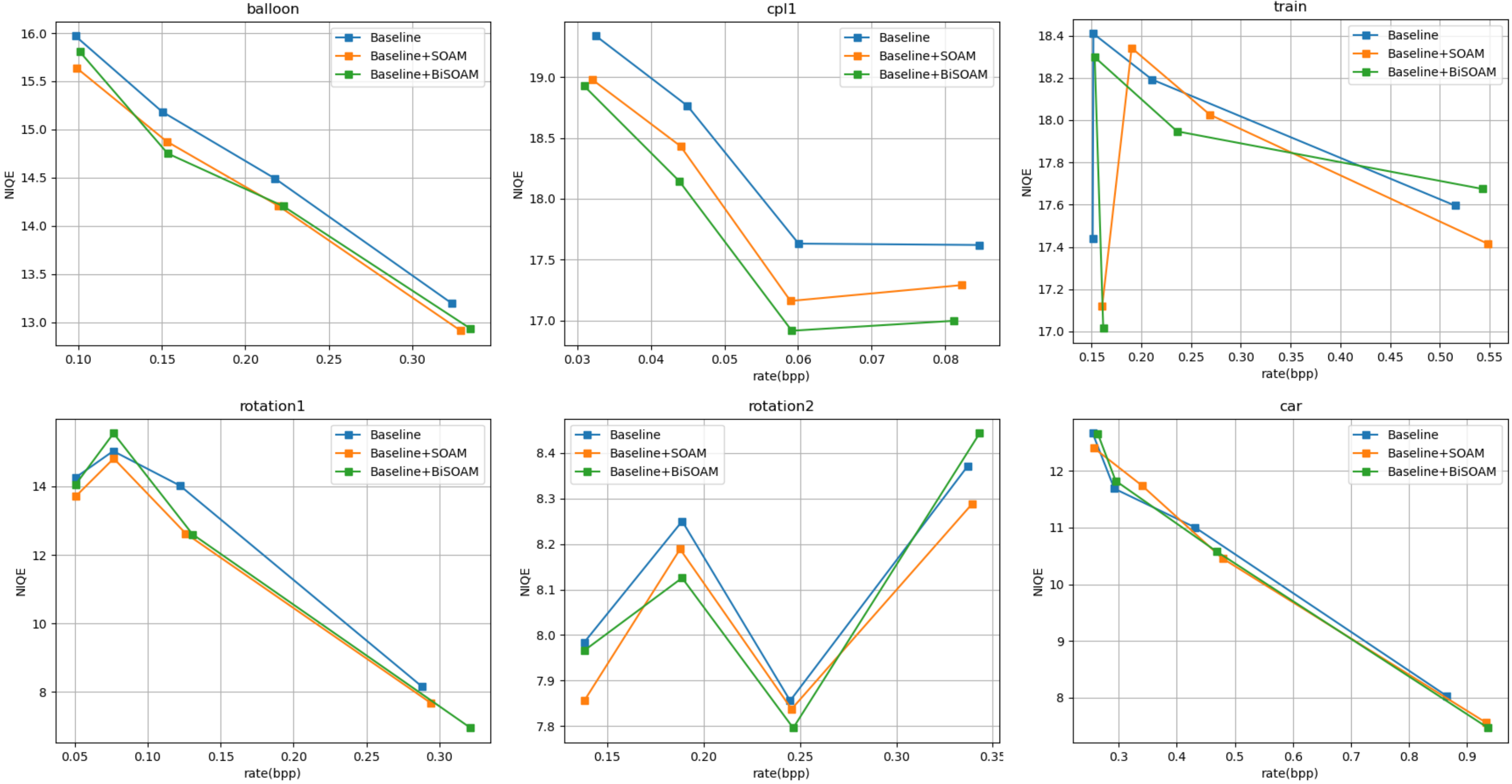}
		\caption{
			NIQE v.s. bit-rate curves on different spike sequences, including \textit{balloon} (top left), \textit{cpl1} (top middle), \textit{train} (top right), \textit{rotation1} (bottom left), \textit{rotation2} (bottom middle) and \textit{car} (bottom right).
			All sequences illustrate that our method outperforms significantly compared with baseline for human perception.
			The BD-rate gain reaches 29.23\% for SOAM and 36.68\% for BiSOAM in average.
		}
		\label{niqe}
	\end{figure*}

	{\bf Non-reference metric based coding performance\quad}
	In the absence of paired data between spike streams and scene images, we rely on non-reference image quality assessment (IQA) methods to evaluate the quality of the reconstructed scene images, taking into account human visual perception. We employ several non-reference IQA metrics for a comprehensive analysis, including (NIQE)~\cite{mittal2012making}, no reference quality metric (NRQM)~\cite{ma2017learning}, neural image assessment (NIMA)~\cite{talebi2018nima}, multi-scale image quality (MUSIQ)~\cite{ke2021musiq}, deep bilinear convolutional neural network (DBCNN)~\cite{zhang2018blind} and perceptual index (PI)~\cite{blau20182018}.
	To provide a meaningful comparison, we select a bit-rate point around 0.3 bpp. The evaluation is performed on six different spike sequences from the PKU-Spike-Recon dataset\footnote{https://www.pkuml.org/resources/pku-spike-recon-dataset.html}, namely \textit{balloon}, \textit{cpl1}, \textit{train}, \textit{rotation1}, \textit{rotation2}, and \textit{car}. The results, shown in Table ~\ref{nfiqa}, highlight the leading performance, which is marked in {\color{red}red}.
	For scenarios with normal-speed, such as \textit{Train}, our method exhibits slightly lower performance compared to the baseline. However, for scenarios involving high-speed motion, the scenes reconstructed using our proposed method demonstrate better quality than those reconstructed using the baseline. This phenomenon can be attributed to the trade-off between temporal and spatial information. Our proposed SOAM and BiSOAM modules prioritize temporal information, possibly resulting in the neglect of certain spatial details. This also explains the performance gap between SOAM and BiSOAM for some high-speed scenarios. When considering sequences with both high and normal-speed motion, our proposed modules consistently deliver more impressive IQA results, demonstrating a significant improvement over the baseline in terms of human perception.
	To provide a visual representation of the NIQE metric versus bit-rate curves for each sequence, please refer to Fig.~\ref{niqe}. On average, our method achieves a BD-rate gain of 29.23\% for SOAM and 36.68\% for BiSOAM, indicating substantial improvements in image quality as assessed by the NIQE metric.
	
	\begin{table*}
		\centering
		\normalsize
		\caption{
			Non-reference metrics comparison between different sequences.
			Among six sequences,    BiSOAM reaches best performance in average for four metrics while SOAM for two.
		}	
		\begin{tabular}{@{}l|ccc|ccc|ccc@{}}
			\hline\hline
			{\bf Metrics} & \multicolumn{3}{c|}{\bf NIQE~\cite{mittal2012making}\enspace($\downarrow$)} & \multicolumn{3}{c|}{\bf NRQM~\cite{ma2017learning}\enspace($\uparrow$)} & \multicolumn{3}{c}{\bf NIMA~\cite{talebi2018nima}\enspace($\uparrow$)} \\
			\hline
			Method & Baseline & \begin{tabular}[c]{@{}l@{}}Baseline\\ +SOAM\end{tabular} &    \begin{tabular}[c]{@{}l@{}}Baseline\\ +BiSOAM\end{tabular} & Baseline & \begin{tabular}[c]{@{}l@{}}Baseline\\ +SOAM\end{tabular} &    \begin{tabular}[c]{@{}l@{}}Baseline\\ +BiSOAM\end{tabular} & Baseline & \begin{tabular}[c]{@{}l@{}}Baseline\\ +SOAM\end{tabular} &    \begin{tabular}[c]{@{}l@{}}Baseline\\ +BiSOAM\end{tabular}\\
			\hline
			balloon & 13.19 &\textbf{12.91} & 12.93 & 5.78 & \textbf{5.84} & 5.83 & 5.89 & 5.90 & \textbf{5.92}  \\
			\hline
			cpl1 & 17.62 & 17.29 & \textbf{17.00} & 2.53 & 2.64 & \textbf{2.68} &	4.36 & \textbf{4.39} & 4.36\\
			\hline
			train & 17.60 & \textbf{17.41} & 17.67 & \textbf{3.16} & 3.08 & 3.09 &	4.02 & \textbf{4.09} & \textbf{4.09} \\
			\hline
			rotation1 & 8.16 & 7.68 & \textbf{6.95} & 3.55 & 3.59 & \textbf{3.78} &	\textbf{4.84} & 4.80 & 4.64 \\
			\hline
			rotation2 & 8.37 & \textbf{8.29} & 8.44 & 7.74 & \textbf{7.75} & \textbf{7.75} &	5.43 & \textbf{5.46} & 5.43 \\
			\hline
			car & 8.02 & 7.56 & \textbf{7.47} & 6.96 & \textbf{6.97} & 6.96 &	4.87 & \textbf{4.89} & 4.87 \\
			\hline
			\textbf{Average} & 12.16 & 11.86 & \textcolor{red}{\textbf{11.74}} & 4.95 & 4.98 & \textcolor{red}{\textbf{5.02}} & 4.9 & \textcolor{red}{\textbf{4.92}} & 4.89\\
			\hline
			\hline
			{\bf Metrics} & \multicolumn{3}{c|}{\bf MUSIQ~\cite{ke2021musiq}\enspace($\uparrow$)} & \multicolumn{3}{c|}{\bf DBCNN~\cite{zhang2018blind}\enspace($\uparrow$)} & \multicolumn{3}{c}{\bf PI~\cite{blau20182018}\enspace($\downarrow$)}  \\
			\hline
			Method & Baseline & \begin{tabular}[c]{@{}l@{}}Baseline\\ +SOAM\end{tabular} &    \begin{tabular}[c]{@{}l@{}}Baseline\\ +BiSOAM\end{tabular} & Baseline & \begin{tabular}[c]{@{}l@{}}Baseline\\ +SOAM\end{tabular} &    \begin{tabular}[c]{@{}l@{}}Baseline\\ +BiSOAM\end{tabular} & Baseline & \begin{tabular}[c]{@{}l@{}}Baseline\\ +SOAM\end{tabular} &    \begin{tabular}[c]{@{}l@{}}Baseline\\ +BiSOAM\end{tabular}\\
			\hline
			balloon &	44.36 & \textbf{44.81} & 44.68 &	\textbf{48.82} & 48.74 & 48.53 &	9.47 & 9.42 & \textbf{9.35}    \\
			\hline
			cpl1 &	29.47 & 29.71 & \textbf{30.00} &	27.99 & 28.16 & \textbf{28.68} & 12.75 & 12.81 & \textbf{12.73}     \\
			\hline
			train &	\textbf{37.93} & 37.14 & 37.45 &	36.34 & 36.02 & 36.50 & \textbf{11.92} & 12.01 & 12.31    \\
			\hline
			rotation1 &	48.81 & 48.66 & \textbf{49.02} &	46.21 & \textbf{47.21} & 45.55 &	7.76 & 7.70 & \textbf{7.29}  \\
			\hline
			rotation2 &	59.95 & 59.49 & \textbf{59.97} &	60.63 & 60.49 & \textbf{60.86} &	5.35 & \textbf{5.30} & 5.36  \\
			\hline
			car &	53.49 & 52.90 & \textbf{53.72} &	51.09 & \textbf{51.14} & 50.93 &	5.19 & 5.16 & \textbf{5.15}\\
			\hline
			\textbf{Average} & 45.67 & 45.45 & \textcolor{red}{\textbf{45.81}} & 45.18 & \textcolor{red}{\textbf{45.29}} & 45.18 & 8.74 & 8.73 & \textcolor{red}{\textbf{8.7}} \\
			\hline\hline
		\end{tabular}
		\label{nfiqa}
	\end{table*}

	\subsection{Compression efficiency in ISI domain}
	In addition to evaluating the compression performance in the scene domain, it is crucial to assess the compression efficiency in the spike domain. The inter-spike interval (ISI) is commonly used as an efficient representation for spike data. To evaluate the fidelity of the reconstructed spike sequences, we calculate the PSNR values between the pristine ISI from uncompressed spikes and the generated ISI. These results are shown in Fig.\ref{psnr} (middle panel).
	We observe that the performance of our method surpasses that of the baseline and demonstrates a more significant improvement compared to the PSNR metric in the image domain. The PSNR gain achieved by our method is 23.39\% and 34.54\% respectively, indicating that the temporal context introduced in the SOAM and BiSOAM modules has a stronger impact on the ISI metric. Similar properties are also observed in other ISI-relevant domains, such as the firing rate shown in Fig.\ref{psnr} (right panel), which measures the fidelity of informative content. The firing rate $Fr$ is calculated solely based on the ISI, and it is evident that our proposed method outperforms both conventional and learned codecs in terms of signal-level fidelity.
	These findings demonstrate the effectiveness of our method in preserving the fidelity of spike-related information during compression, making it a superior choice for spike data compression compared to state-of-the-art conventional and learned codecs.
	
	The process of spike generation and ISI calculation is non-differentiable, making it challenging to include them in end-to-end training. However, the optimization for the firing rate can be treated similarly to that for scene images. According to Equation \ref{ISI-luminance}, the mathematical expectation of the firing rate is related to the mathematical expectation of the luminance, and it can be formulated as follows:
	\begin{equation}
		\mathbb{E}[Fr_n] = \frac{1}{\mathbb{E}[T^{ISI}_n]} = \frac{\alpha}{\theta}\mathbb{E}[I_n].
	\end{equation}
	You are correct. Since the firing rate is linearly related to the luminance with a constant factor $\frac{\alpha}{\theta}$, optimizing the luminance is equivalent to optimizing the firing rate. The predicted firing rate will be closest to the actual firing rate if and only if the predicted luminance is closest to the actual luminance.
	In the proposed framework, the first two stages involve joint training of the scene reconstruction and image compression modules. By optimizing these two modules, we indirectly optimize the luminance and firing rate. Therefore, reaching an optimal solution for the first two stages of the framework also implies reaching an optimal solution for the third stage.
	To further improve the framework, it is indeed possible to consider joint optimization of all three stages. This can be achieved by incorporating operations such as quantization, which are commonly used in learning-based codecs. By jointly optimizing the compression performance in both the scene and spike domains, the framework can potentially achieve even better results. Further research in this direction would be valuable for enhancing the overall performance of the proposed framework.
	
	\subsection{Comparison of model complexity}
	The comparison of trainable parameters and floating point operations (FLOPs) between the baseline model and the proposed methods, SOAM and BiSOAM, is presented in Table~\ref{flops}. The results show that the introduction of SOAM incurs an additional 0.23\% trainable parameters and 1.07\% computational complexity, resulting in a 2.31\% gain on BD-rate. Similarly, the introduction of BiSOAM leads to a 0.55\% increase in trainable parameters and 2.64\% increase in computational complexity, resulting in a 2.53\% gain on BD-rate.
	These findings demonstrate that the spike-oriented attention modules, namely SOAM and BiSOAM, offer significant improvements in spike compression performance with minimal additional cost. Despite the slight increase in trainable parameters and computational complexity, the benefits in terms of BD-rate gain highlight the effectiveness of integrating these attention modules into the framework. Overall, the introduction of these modules enhances the compression efficiency of the framework, making it a more powerful and efficient solution for spike compression tasks.
	
	\begin{table}
		\centering
		\caption{
			Comparison of parameters and computational complexity between baseline model and our proposed methods.
			Results show that the spike oriented attention modules improve performance of spike compression effectively with insignificant extra cost.
		}	
		\begin{tabular}{@{}c|c|c|c@{}}
			\hline\hline
			& \textbf{Baseline} & \textbf{Baseline + SOAM} & \textbf{Baseline + BiSOAM}\\
			\hline
			\textbf{Params (M)} & 48.72 & 48.82 & 48.98 \\
			\hline
			\textbf{FLOPs (G)} & 685.52 & 692.88 & 703.60 \\
			\hline\hline
		\end{tabular}
		\label{flops}
	\end{table}

	\subsection{Comparison on downstream application}
\begin{figure}
		\centering
		\includegraphics[width=.9\linewidth]{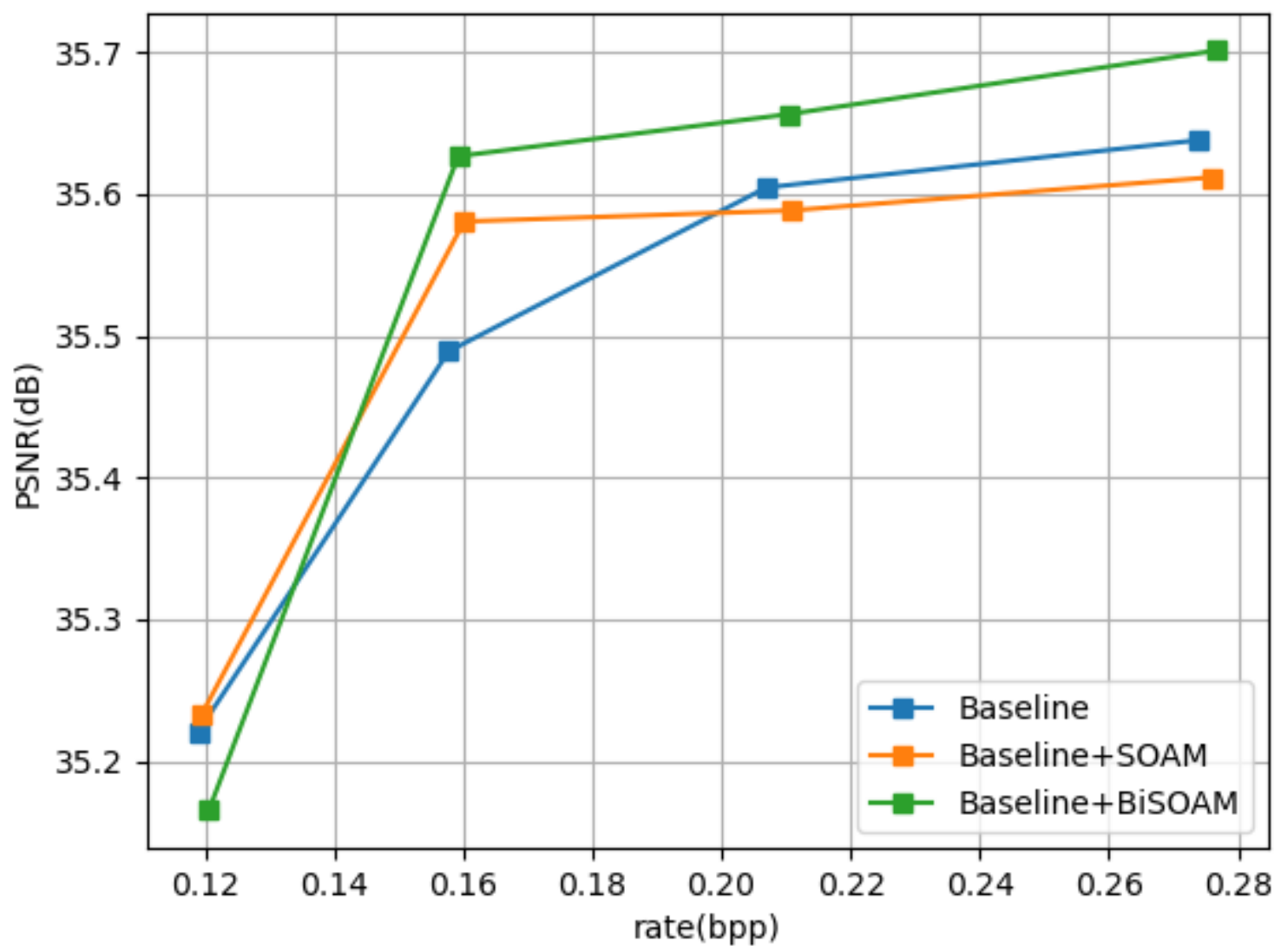}
		\caption{
			PSNR v.s. bpp curves of baseline and proposed method.
			PSNR is calculated between scenes reconstructed from raw and decoded spike stream.
			Results show our methods have great informative fidelity compared with baseline model, 
		}
		\label{psnr_regenerate}
	\end{figure}

	We performed a comparison between scenes reconstructed from raw and decoded spike streams to evaluate the level of informative fidelity, as shown in Fig.~\ref{psnr_regenerate}. For the reconstruction, we employed the method proposed in~\cite{zhao2021spk2imgnet}. It is noteworthy that the scenes reconstructed from the decoded spike stream exhibit a striking similarity to those reconstructed from the raw stream, especially in scenarios with high bit-rates. This observation suggests that the information contained in both the raw and decoded spike streams is roughly equivalent, thus validating the suitability of the proposed scene reconstruction-based framework for spike data compression with a focus on preserving informative fidelity.
	Furthermore, the proposed SOAM and BiSOAM techniques effectively capture implicit features present in the spike streams, resulting in the preservation of more information at the same bit-rate. This indicates that these attention modules enhance the compression efficiency by effectively utilizing the available information in the spike streams. The visual similarity between the reconstructed scenes further supports the notion that the proposed framework, along with the integration of SOAM and BiSOAM, successfully preserves informative fidelity in the compressed spike data.
	
	\section{Conclusion}
	
	In this paper, we propose a learning based compression framework for spike data with  scene reconstruction and coding, which is the first learned framework aiming to encoding the continuous spike streams.
	We propose a compression-friendly representation of spike data to compress efficiently with assistance of spike oriented attention mechanism.
	Extensive experimental results show that this scene reconstruction based method can compress spike sequences with ultra-high temporal resolution efficiently and reconstruct with low distortion, which is significantly better than the SOTA learning based codec.
	This provides a strong baseline for learning-based spike stream compression task and facilitates future study. 
	
	\bibliographystyle{unsrt}
	\bibliography{tpami}
	
\end{document}